\documentclass[10pt,twocolumn,letterpaper]{article}
\usepackage{pifont}
\usepackage{threeparttable}
\usepackage{iccv}             
\usepackage[accsupp]{axessibility}

\definecolor{iccvblue}{rgb}{0.21,0.49,0.74}
\usepackage[pagebackref,breaklinks,colorlinks,allcolors=iccvblue]{hyperref}
\usepackage[all]{hypcap}
\def\paperID{14342}
\def\confName{ICCV}
\def\confYear{2025}

\title{VideoMiner: Iteratively Grounding Key Frames of Hour-Long Videos via Tree-based Group Relative Policy Optimization}

\author{Xinye Cao\textsuperscript{\rm 1}\footnotemark[1]
\and
Hongcan Guo\textsuperscript{\rm 1}\footnotemark[1]\thanks{Equal contribution}
\and
Jiawen Qian\textsuperscript{\rm 1}\footnotemark[1]
\and
Guoshun Nan\textsuperscript{\rm 1}\footnotemark[2]\thanks{Corresponding author}
\and
Chao Wang\textsuperscript{\rm 1}
\and
Yuqi Pan\textsuperscript{\rm 1}
\and
Tianhao Hou\textsuperscript{\rm 1}
\and
Xiaojuan Wang\textsuperscript{\rm 1}
\and
Yutong Gao\textsuperscript{\rm 2}
\and
\\
\textsuperscript{\rm 1}Beijing University of Posts and Telecommunications, China\\
\textsuperscript{\rm 2}Minzu University of China, China\\
{\tt\small \{caoxinye, ai.guohc, qjwww, nanguo2021, wangchao0317, panyuqi2022, datoucai,}\\
{\tt\small wj2718\}@bupt.edu.cn, ytgao92@muc.edu.cn}
}

\makeatletter
\AtBeginDocument{
\let\@oldmaketitle\@maketitle
\renewcommand{\@maketitle}{\@oldmaketitle
  \vspace{-25pt}
 
  \includegraphics[width=1\textwidth, trim= 22 30 25 18, clip]{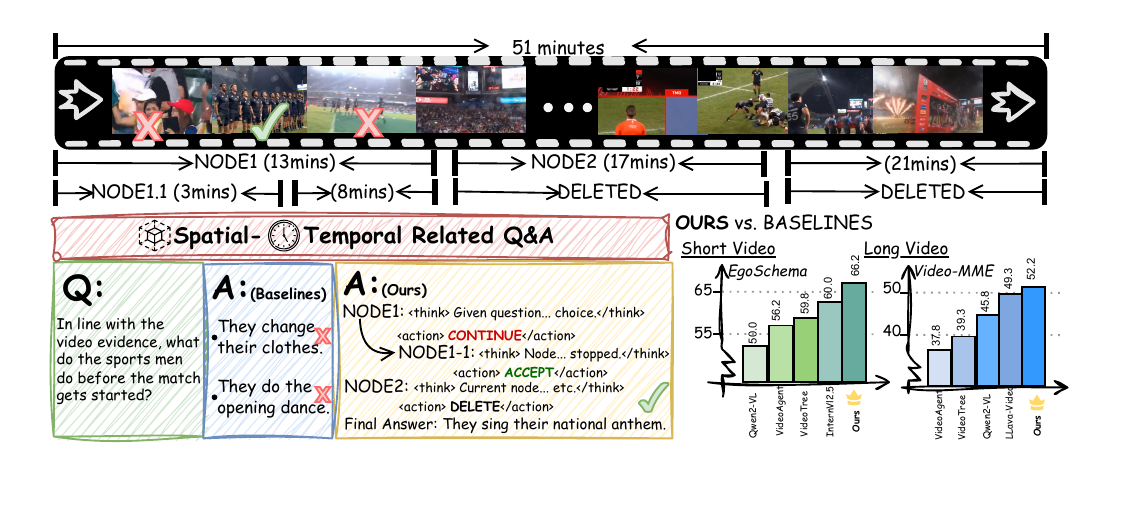}
  \vspace{-10pt}
  \captionof{figure}{Illustration of spatial-temporal related Q\&A performance on long videos. The input is a 51-minute video with a question about athletes' actions before the match. Baselines provide answers such as changing clothes or dancing. Our method, incentivizing chain-of-thought ability by reinforcement learning, correctly identifies the act of singing the national anthem by locating key frames. The right side of the figure shows our superior performance against multiple baselines across both short and long videos.}
  \label{fig:teaser}
  \vspace{16pt}
 }
}
\makeatother
\usepackage[table]{xcolor}
\usepackage{colortbl}      
\usepackage{multirow}   
\usepackage{adjustbox}

\begin{document}
\maketitle
\begin{abstract}
Understanding hour-long videos with multi-modal large language models (MM-LLMs) enriches the landscape of human-centered AI applications. However, for end-to-end video understanding with LLMs, uniformly sampling video frames results in LLMs being overwhelmed by a vast amount of irrelevant information as video length increases. Existing hierarchical key frame extraction methods improve the accuracy of video understanding but still face two critical challenges. 1) How can the interference of extensive redundant information in long videos be mitigated? 2) How can a model dynamically adapt to complex hierarchical structures while accurately identifying key frames? To address these issues, we propose VideoMiner, which iteratively segments, captions, and clusters long videos, forming a hierarchical tree structure. The proposed VideoMiner progresses from long videos to events to frames while preserving temporal coherence, effectively addressing the first challenge. To precisely locate key frames, we introduce T-GRPO, a tree-based group relative policy optimization in reinforcement learning method that guides the exploration of the VideoMiner. The proposed T-GRPO is specifically designed for tree structures, integrating spatiotemporal information at the event level while being guided by the question, thus solving the second challenge. We achieve superior performance in all long-video understanding tasks and uncover several interesting insights. Our proposed T-GRPO surprisingly incentivizes the model to spontaneously generate a reasoning chain. Additionally, the designed tree growth auxin dynamically adjusts the expansion depth, obtaining accuracy and efficiency gains. The code is publicly available at \url{https://github.com/caoxinye/VideoMiner}.
\end{abstract}    
\vspace{-7mm}
\section{Introduction}
\label{sec:intro}
MM-LLMs \cite{DBLP:conf/cvpr/RenYL0H24,DBLP:conf/acl/CaffagniCBMS0CC24,DBLP:conf/icml/Wu0Q0C24, 10654917, du2024docmsu}, which integrate LLMs \cite{DBLP:conf/emnlp/ZhangLB23,DBLP:conf/cvpr/ChenLWLSGLGMS24} with vision encoders \cite{DBLP:conf/icml/0003GYZYSFQW0HS24}, extend their inherent ability to comprehend human-like text to encompass advanced visual reasoning tasks. Given the heterogeneity of visual inputs, MM-LLMs exhibit variations in model design and training to understand images, short videos, and long videos. Long video understanding \cite{DBLP:conf/eccv/WangZZY24,DBLP:conf/nips/QianDZZDLW24} capability of MM-LLMs enriches the landscape of human-centered AI applications \cite{DBLP:journals/pami/RongLNFQUSKK24, nan2020reasoning}, including automatic detection of highlight moments in sports footage, summarization of cinematic narratives, and anomaly detection in surveillance videos \cite{DBLP:conf/cvpr/DoshiY20}.

However, unlike static images and short videos, long videos typically consist of thousands of frames and span hours, presenting richer spatial detail and more intricate temporal dynamics \cite{DBLP:conf/cvpr/AafaqALGM19}. MM-LLMs struggle to ground key frames among massive irrelevant information as the video length increases. This leads to the first challenge: 1) how can the interference of extensive redundant information in long videos be mitigated? Hierarchical key frame extraction facilitates MM-LLMs in understanding long videos but may disrupt the original video structure, leading to the loss of temporal information \cite{DBLP:conf/cvpr/BuchEG00N22}. Key frame extraction methods \cite{DBLP:conf/acl/0001RKK24} need to effectively integrate with the hierarchical structure while incorporating multi-level information. Therefore, we introduce the second challenge: 2) how can a model dynamically adapt to complex hierarchical structures \cite{DBLP:conf/cvpr/YeLQWH022,DBLP:conf/cvpr/IslamHYNTB24} while accurately identifying key frames?

Existing LLM-based approaches for long video understanding include end-to-end \cite{DBLP:conf/emnlp/0010LIWYBB24,DBLP:conf/cvpr/LinLL0G0LW22} and hierarchical structure \cite{DBLP:journals/tip/GaoLZSWS22}. For end-to-end structure, video content is typically simplified into a flat list of subtitles, leading to irrelevant information that increases exponentially as the video length extends. In contrast, hierarchical video representations introduce some level of structure to reduce the complexity of long videos. The most relevant work is newly emerged VideoTree \cite{DBLP:journals/corr/abs-2405-19209}, including visual clustering, frame caption \cite{DBLP:conf/eccv/ChenWZH18}, and correlation scoring. However, it is difficult to effectively extract key frames in hour-long videos, thereby hindering long video understanding of MM-LLMs. As illustrated in Figure \ref{fig:teaser}, MM-LLMs tend to be influenced by irrelevant frames, leading to incorrect responses.

To address the aforementioned challenges, we propose VideoMiner, a novel reinforcement learning-based video understanding tree. To preserve the temporal structure of long videos, we segment the video based on dynamic events and then cluster the captions, with each clustered event forming a tree node. VideoMiner constructs a hierarchical tree structure that progresses from coarse to fine granularity, transitioning from the video level to events, and then to frames while maintaining temporal coherence.  

For key frame extraction, we establish three guiding principles: 1) integrating spatial-temporal information at the event level, 2) ensuring query-oriented exploration, and 3) adapting to the hierarchical tree structure. Based on the three principles, we propose T-GRPO, which dynamically determines key frame exploration based on event captions, question inputs, and node depth. To efficiently search for key frames, we introduce a tree growth rate mechanism to regulate exploration depth.  

As illustrated in Figure \ref{fig:teaser}, our method significantly outperforms other baselines on both long-video and short-video benchmarks, boosting the long video understanding of MM-LLMs. Furthermore, the policy model trained with our proposed T-GRPO spontaneously develops a reasoning chain to generate in-depth responses. The main contributions of this paper are as follows.
\begin{itemize}
    \item We propose VideoMiner, an adaptive tree structure that decomposes long videos into a hierarchical set of events while preserving temporal coherence, facilitating efficient key frame grounding.
    \item We propose T-GRPO, a tree-based group relative policy optimization for reinforcement learning, adaptively exploring key frames in VideoMiner.
    \item We conduct extensive experiments on four well-known benchmarks against ten baselines, proving the superiority of our methods. Ablation studies confirm the effectiveness of the clustering and T-GRPO methods. Interestingly, training with T-GRPO invokes the model's reasoning chains, guiding to in-depth inference.
\end{itemize}

\section{Related Work}
\label{sec:formatting}

\paragraph{Long Video Understanding with LLMs.} 
Recent works \citep{DBLP:conf/emnlp/0010LIWYBB24,DBLP:conf/cvpr/SongCWZZWCG0ZLH24,DBLP:conf/eccv/LiuLTGSL24,DBLP:conf/nips/QianDZZDLW24} have expanded the capabilities of LLMs to video understanding. For end-to-end video understanding, Video-LLaVA \cite{DBLP:conf/emnlp/LinYZCNJ024}, LLaVA-Video \cite{DBLP:journals/corr/abs-2410-02713}, InternVL2.5 \cite{DBLP:journals/corr/abs-2412-05271} and Qwen2-VL \cite{DBLP:journals/corr/abs-2409-12191} propose a language-guided video understanding method. 
For hierarchical video understanding, VideoAgent \cite{DBLP:conf/eccv/WangZZY24} leverages an LLM agent to conduct multi-round frame searches. The most closely related VideoTree \cite{DBLP:journals/corr/abs-2405-19209} dynamically extracts query-relevant keyframes with tree structure. LLoVi \cite{DBLP:conf/emnlp/0010LIWYBB24} introduces a long-video QA framework that segments videos into clips, generates textual descriptions for each segment, and then applies LLM-based reasoning and multi-round summarization to enhance QA performance. Different from previous works, we propose VideoMiner, which forms a tree structure to extract key frames of long videos while preserving temporal relations. The tree‐exploration process is adaptively controlled by T-GRPO, integrating fine-grained spatiotemporal details.

\begin{figure*}[ht]
\vskip 0.2in
\begin{center}
\centerline{\includegraphics[width=1\textwidth,trim=25 10 20 10,clip]{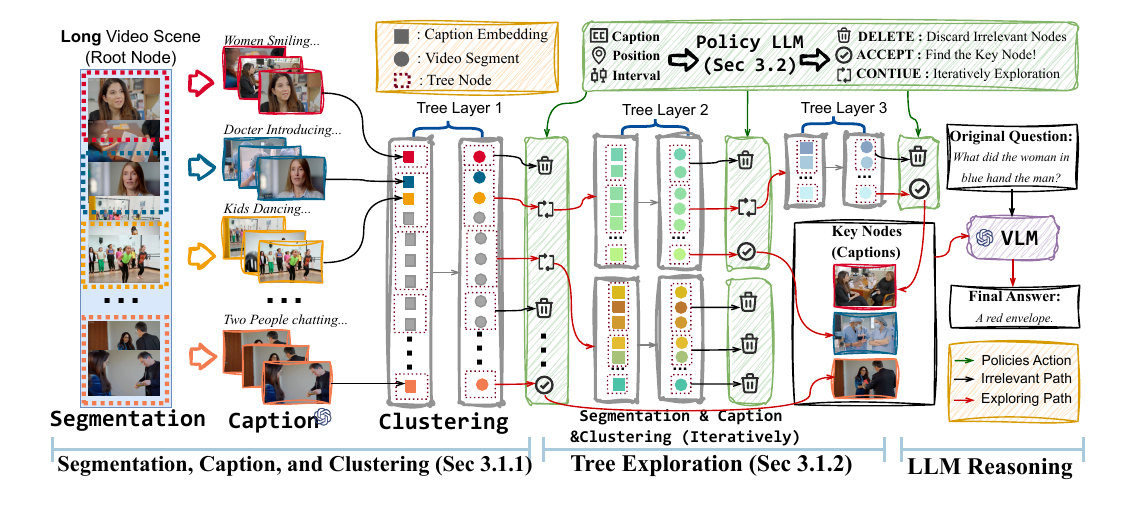}}
\caption{Illustration of the workflow of our proposed VideoMiner. The long video undergoes iterative segmentation, captioning, and clustering to construct a hierarchical tree structure. The policy model governs the exploration of tree nodes and identifies key frames. The selected key frames, along with the original question, are then fed into the VLM for long-video reasoning, producing the final answer.}
\label{fig:overview}
\end{center}
\vspace{-12mm}
\end{figure*}

\vspace{-4mm}
\paragraph{Reinforcement Learning in Video Grounding.}
Reinforcement learning has been widely applied to video-related tasks, such as video summarization \citep{Zhou_Qiao_Xiang_2018,WOS:000750373700004}, action recognition \citep{WOS:000457843605049,WOS:000432398300016,Dong_Zhang_Tan_2019}, captioning \citep{WOS:000457843604038,WOS:000587912800009,WOS:000761218500002},  representation learning \citep{WOS:000966336700001,WOS:001397798100003}, and grounding \citep{DBLP:conf/cvpr/ZengXHCTG20,DBLP:conf/cvpr/NanQXLLZL21,DBLP:conf/cvpr/YangMSLS22}. 
For video grounding, RWM-RL \cite{He_Zhao_Huang_Li_Liu_Wen_2019} formulates the task as a sequential decision-making problem by learning an agent which regulates the boundaries of temporal grounding. 
The most relevant work \cite{Wu_Li_Liu_Lin_2020} designs a tree-structured policy-based progressive reinforcement learning framework to sequentially regulate the temporal boundary. Different from the above methods, our proposed T-GRPO refines the GRPO \cite{DBLP:journals/corr/abs-2402-03300} method via a tree structure, tailored to long video understanding tasks.

\section{Method}
In this section, we first introduce the overall process of our proposed VideoMiner, as described in the \S\ref{subsec: Workflow of the Proposed VideoMiner}.
Then we give a detailed procedure of the proposed T-GRPO, as elaborated in the subsequent \S\ref{TGRPO}. The details of the methods and reasoning process are provided in Appendix A of the supplementary material.

\subsection{Workflow of the Proposed VideoMiner} \label{subsec: Workflow of the Proposed VideoMiner}
As illustrated in Figure \ref{fig:overview}, the proposed VideoMiner basically consists of three components: scene segmentation and caption, T-GRPO based tree exploration, and LLM reasoning. 
The input long video is temporally segmented into events, which are then processed by a VLM (Vision Language Model) to generate captions based on the given question. 
We then perform clustering based on captions, where each cluster is treated as a tree node. 
The policy model in T-GRPO determines whether a node should continue expanding. 
If further expansion is required, the node undergoes an iterative process of segmentation, captions generation, and clustering to create new child nodes. 
This process continues until the policy model identifies all key frames. 
Finally, captions of key frames and the original question are fed into VLM to perform reasoning and give the final answer.
\subsubsection{Segmentation, Caption, and Clustering}
Hour-long videos contain a vast amount of redundant information that is unrelated to the given question. 
To mitigate the complexity of long videos and form a hierarchical structure, we first apply uniform sampling and segment the video based on distinct scenes. 
By adaptively segmenting the video at the event level rather than using discrete frames, we effectively preserve temporal coherence, minimizing the disruption of temporal information during both the segmentation and subsequent clustering processes. 
We formulate the complete process below.

\vspace{-5mm}
\paragraph{Scene Segmentation.}
A long video, after uniform sampling into \( N \) frames, can be represented as a set \( \mathcal{F}_i = \{f_1, \dots, f_t, \dots, f_N\} \). Each frame \( f_t \) is represented by a normalized grayscale histogram, capturing the distribution of intensity levels within the image: 
\begin{align}
& H_t(k) = \frac{1}{WH} \sum_{i=1}^{W} \sum_{j=1}^{H} \delta(\text{gray}(f_t(i,j)) - k), \notag
\\& \quad k \in \{0,1,\dots,255\},
\end{align}
where \( H_t(k) \) denotes the normalized histogram value at grayscale level \( k \), \( W \times H \) is the image resolution, and \( \text{gray}(f_t(i,j)) \) represents the grayscale intensity at coordinate \( (i,j) \) in frame \( t \). The Kronecker delta function $\delta(x) = 1$ only when \( x = 0 \).

To quantify frame-to-frame similarity, we employ the Bhattacharyya distance \( D_i \) between consecutive histogram distributions, constructing a similarity sequence as follows:
\begin{equation}
    D_i = -\ln \sum_{k=0}^{255} \sqrt{H_i(k)H_{i+1}(k)},
\end{equation}
where $H_i(k)$ and $H_{i+1}(k)$ represent the normalized grayscale histograms of frames i and i+1, respectively.

The segmentation threshold \( \tau \) is determined by selecting the top \( K-1 \) largest change points. The corresponding segmentation points \( \{p_1, \dots, p_{K-1}\} \) are identified, resulting in the final scene partitioning:  
\begin{equation}
    E_m = \left\{ 
    \begin{array}{cl}
    \{f_1,...,f_{p_1}\} & m=1 \\
    \{f_{p_{m-1}+1},...,f_{p_m}\} & 2 \leq m \leq K-1 \\
    \{f_{p_{K-1}+1},...,f_N\} & m=K 
    \end{array} ,
    \right.
\end{equation}
after scene segmentation, the input long video $\mathcal{F}_i$ is partitioned into \( K \) distinct scenes $E = \{E_1, \dots, E_K\} $.
\vspace{-5mm}
\paragraph{Caption Generation.}
Each event contains a continuous sequence of frames. To capture specific information relevant to the user-provided question \( Q \) and improve the clustering efficiency, a VLM is utilized to generate captions for each event. The captions for the \( m \)-th event is defined as: 
\begin{equation}
\text{Caption}_m = \mathrm{VLM}(E_m, Q), \quad m=1, \dots, K.
\vspace{-3mm}
\end{equation}

\paragraph{Clustering.}
To transform a long video into a hierarchical tree structure, we cluster events based on captions to form tree nodes. Each textual description \( \text{caption}_i \) is mapped to a vector representation using an embedding model:
\begin{equation}
{v}_m = \text{Embedding}(\text{Caption}_m),
\end{equation}
where the extracted embeddings form a feature matrix \( {V} \in \mathbb{R}^{K \times d} \). Next, a density-based clustering algorithm, DBSCAN, is applied to group the \( K \) events into \( C \) semantic events with similar spatial characteristics:
\begin{equation}
\{v_1,...,v_K\} \xrightarrow[{\epsilon},\text{minPts}]{\text{DBSCAN}}\{l_1,...,l_C\},
\end{equation}
where \( l_i \) denotes the cluster label assigned to the $i$-th sub-scene through clustering, and \( \epsilon \) represents the neighborhood radius while \( \text{minPts} \) denotes the minimum density threshold. The final event segmentation corresponds to the $i$-th node \( N_{i} \), which is associated with the label \( l_i \). The number of resulting clusters satisfies \( C \leq K \), ensuring that semantically related scenes are grouped together to form higher-level structural nodes within the tree.

\subsubsection{Tree Exploration}
After segmentation, caption, and clustering to form tree nodes \( N \), policy model in our proposed T-GRPO decides which nodes can iteratively expand into new nodes. As the tree grows, the long video is decomposed into a hierarchical structure, progressing from coarse to fine, where a deeper layer of the tree contains more fine-grained information. The action of the policy model includes three states: accept, continue, and delete. Specifically, accept indicates that the node contains sufficient key frames to answer the question, requiring no further exploration. Continue suggests that the node may be relevant to the query and should be further expanded to new nodes for deeper exploration. Delete signifies that the node is irrelevant to the question and can be discarded without further expansion.

As the core component, the policy model \( PM \) determines the tree growth process, which is designed based on three aspects: spatiotemporal information integration, question orientation, and structural adaptability. Following the three design principles, the policy model takes three inputs: event captions \( {Caption}_m \), the user question \( Q \), and node depth \( depth(N_i) \). The event captions preserve the temporal continuity of the original long video, while the question-driven captions reflect spatial information. Incorporating the question ensures that the model’s decision-making remains closely aligned with the user's intention. Node depth provides positional information within the hierarchical structure. In Section \ref{TGRPO}, we introduce the concept of tree auxin to regulate excessive exploration, thereby enhancing localization accuracy and efficiency. The output of the policy model \( State(N_i) \) can be represented as:
\begin{equation}
State(N_i) = \mathrm{PM}\left({Caption}_m, Q, depth(N_i) \right).
\end{equation}
All nodes with the state of accept represent the selected key frames. These key frames are collected, along with the user's question, are fed into the VLM for inference to generate the final result.

\begin{figure*}[ht]
\begin{center}
\centerline{\includegraphics[width=0.9\textwidth,trim=10 120 15 0,clip]{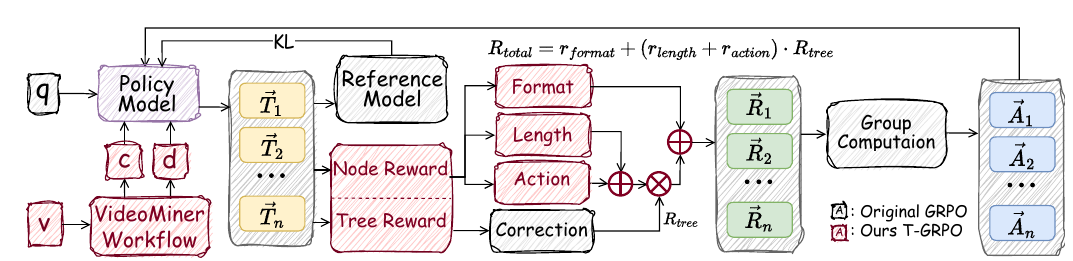}}
\caption{Illustration of the proposed T-GRPO. To highlight the differences from GRPO, we visualize the original GRPO components in gray, while newly introduced components are marked in red. Unlike GRPO, which primarily optimizes the final output, our approach focuses on the tree generation process, including node exploration behavior. To adapt to the hierarchical structure and video understanding tasks, we modify the tree framework and redesign the reward function accordingly.}
\vspace{-3mm}
\label{fig:TGRPO}
\end{center}
\end{figure*}

\subsection{Tree-Based Group Relative Policy Optimization}
\label{TGRPO}
GRPO \citep{DBLP:journals/corr/abs-2402-03300} eliminates the need for additional value function approximation as required in PPO (Proximal Policy Optimization) and instead utilizes the average reward from multiple sampled outputs as a baseline, significantly reducing training resource consumption. Building on the concepts of GRPO, we propose T-GRPO, which differs primarily in its adaptation to the tree structure and reward design. 

As illustrated in Figure \ref{fig:TGRPO}, one distinction is the structural adaptation to the tree structure. The policy model takes as input not only the query \( q \) but also the caption and tree depth, producing multiple trees, each containing several nodes. Another distinction lies in the reward function design. To accommodate the unique structure and characteristics of the video understanding tree, we decompose the original reward function into node-level rewards and tree-level rewards, which correspond to intermediate node outputs and the final tree output, respectively.
We provide a detailed explanation of the rollout process, reward design, and loss function formulation for T-GRPO as below.
\vspace{-5mm}
\paragraph{Rollout Process.}
As illustrated in Figure \ref{fig:TGRPO}, we first employ the proposed VideoMiner process to perform a rollout, generating \( n \) distinct trees $T = \{\vec{T_1}, \dots, \vec{T_i}, \dots, \vec{T_n}\}$. The \( i \)-th tree $\vec{T_i} = \{ O_{i1}, \dots, O_{ij}, \dots, O_{iG_i}\}$, where $G_i$ is the number of nodes in $T_i$. \( O_{ij} \) denotes the output of the \( j \)-th node in tree \( T_i \), representing the policy model's decision on whether the node qualifies as a key frame. From \( O_{ij} \), we can extract output format $f_o$, complement length $l_o$ and action decisions $a_o$.
\vspace{-5mm}
\paragraph{Reward Design.}
To guide the policy model in making more structured, detailed, and accurate key-frame decisions, we design two types of rewards for each node. The first is the node-level reward \( R_{node} \), which evaluates the quality of individual node decisions, while the second is the tree-level reward \( R_{tree} \), which reflects the correctness of the final tree-level outcome. The node-level reward \( R_{node} \) is further divided into three components: a format reward, which is independent of the final output but ensures structural consistency, and a length and action reward, which directly impacts the accuracy of the final result.

The format reward can be expressed as:
\begin{equation}
\label{eq:format}
r_{\text{format}}(f_o) = \delta_{max} \cdot \mathbb{I}_{\text{max}} + \delta_{\mathrm{corr}} \cdot  \mathbb{I}_{\text{corr}},
\end{equation}
where $\mathbb{I}$ is indicate function. The max condition indicates full compliance with the format, corresponding to a reward of \( \delta_{max} \). The corr condition signifies partial compliance, where the format is still sufficient for correct extraction, corresponding to a reward of \( \delta_{corr} \).

The completion length reward can be expressed as:
\begin{equation}\label{eq:length}
 r_{\text{length}}(l_o) = \rho \exp\!\Biggl(-\frac{(l_o - l_t)^2}{2\sigma^2}\Biggr),
\end{equation}
here, \( l_o \) represents the length of the generated response in tokens, while \( l_t \) denotes the target token length. The parameter \( \sigma \) controls the smoothness of the reward curve and \( \rho \) is a scaling factor. By modeling the reward with a Gaussian distribution, we effectively regulate the target token length of the response. Empirically, we observed that increasing the response length improves overall performance.

The action reward can be expressed as:
\begin{align} \label{eq:action}
r_{\text{action}}(a_o) = & \delta_d\,\mathbb{I}_{\{\text{"delete"}\in a_o\}} + \delta_a\,\mathbb{I}_{\{\text{"accept"}\in a_o\}} + \notag\\&
\delta_c\,\mathbb{I}_{\{\text{"continue"}\in a_o\}},
\end{align}
\begin{equation}
\lambda_{auxin} = \frac{\delta_d + \delta_a}{2\delta_c},
\end{equation}
here, \( \delta_d \), \( \delta_a \), and \( \delta_c \) represent the reward values assigned to the detected states "delete," "accept," and "continue," respectively. The reward for the delete state is the highest, followed by accept, which is slightly lower, and continue, which receives the lowest reward among the three. Inspired by the auxin of plants, we define \( \lambda_{auxin} \) to adaptively regulate tree expansion. By moderating the growth of the tree to a certain extent, we can enhance localization efficiency.

Among the three reward components, \( r_{\text{length}} \) and \( r_{\text{action}} \) directly impact the effectiveness of the final decision. Therefore, we compute the total reward for the policy model output using the following equation:
\begin{equation}
R_{total} = r_{format} + (r_{length} + r_{action}) \cdot R_{tree}.
\end{equation}
This design ensures that the model considers both the correctness of the final decision and the control of response length and action selection. By adjusting the growth factor $\lambda_{auxin}$, the model is encouraged to prefer the accept and delete actions when appropriate, thereby improving efficiency while maintaining decision accuracy.
\vspace{-4.8mm}
\paragraph{Loss Function.}
The total reward $r_{ij}$ for each node is used to compute the group advantage, which quantifies the advantage of each node within the hierarchical structure.
\begin{equation}\label{eq:advantage}
A_{ij} = \frac{r_{ij} - \text{mean}(\{r_{11}, r_{12}, \cdots, r_{nG_n}\})}{\text{std}(\{r_{11}, r_{12}, \cdots, r_{nG_n}\})}.
\end{equation}
Finally, the policy model is updated using a loss function tailored to tree structure optimizing its decision-making process.
\begin{multline} \label{eq:loss}
\mathcal{J}_{T-GRPO}(\theta) \\
= \mathbb{E}[q \sim P(Q), \{o_{ij}\}_{\substack{i=1,\ldots,G \\ j=1,\ldots,N_i}} \sim \pi_{\theta_{old}}(O|q)] \\
\Bigg[
\frac{1}{\sum_{i=1}^n G_i} \sum_{i=1}^n \sum_{j=1}^{G_i} \Bigg(
Adv_{ij}
- \beta \mathbb{D}_{KL} (\pi_\theta || \pi_{ref}) 
\Bigg)\Bigg],
\end{multline}
here, \( q \) represents the current input, and \( o_{ij} \) denotes the corresponding output. The policies \( \pi_{\theta} \), \( \pi_{\theta_{old}} \), and \( \pi_{\theta_{ref}} \) correspond to the current model, the policy from the previous step, and the reference model, respectively. The hyperparameters \( \epsilon \) and \( \beta \) control the clipping coefficient and the KL constraint strength, respectively. The \( Adv_{ij} \) is the product of advantage and policy probability ratio,
\begin{multline} 
\vspace{-1mm}
\label{eq:adv_ratio}
Adv_{ij}
= 
\min \Bigg( 
\frac{\pi_\theta(o_{ij}|q)}{\pi_{\theta_{old}}(o_{ij}|q)} A_{ij},  \\
\text{clip} \left( 
\frac{\pi_\theta(o_{ij}|q)}{\pi_{\theta_{old}}(o_{ij}|q)},
1-\epsilon, 1+\epsilon 
\right) A_{ij} 
\Bigg),
\end{multline}
here, $\text{clip}(\cdot)$ is a clipping function used to constrain policy updates and prevent policy collapse. The policy model updates its parameters based on this loss function, integrating both the global tree structure and individual node outputs, thereby enhancing its reasoning and inference capabilities.

\section{Experiments}

\subsection{Experimental Setup}

\paragraph{Environment.} All experiments are conducted on a server running CentOS Linux 7 (Core) with PyTorch 2.3. The hardware configuration includes 240GB of RAM, a 16-core Intel Xeon CPU, and two NVIDIA A800 GPUs, each equipped with 80GB of memory. 
\vspace{-5mm}
\paragraph{Tasks \& Datasets.} 
We first trained the policy model using the T-GRPO reinforcement learning method on a small-scale subset curated from the well-known open-source Video Question Answering dataset LLaVA-Video-178K \cite{DBLP:journals/corr/abs-2410-02713}. Subsequently, we evaluated VideoMiner across comprehensive video understanding benchmarks covering both long-form and short-term video comprehension. Specifically, EgoSchema \cite{DBLP:conf/nips/MangalamAM23} and MLVU \cite{DBLP:journals/corr/abs-2406-04264} focus exclusively on long-form video understanding, with MLVU extending to hour-level durations. Meanwhile, Video-MME \cite{DBLP:journals/corr/abs-2405-21075} and LongVideoBench \cite{DBLP:conf/nips/WuLCL24} encompass videos spanning multiple granularities: from short clips (tens of seconds to minutes) to extended long-form content (tens of minutes to hours).

\vspace{-5mm}
\noindent
\paragraph{Baselines.} We conduct extensive experiments across multiple strong foundation models. Open-source implementations include Qwen2-VL\cite{DBLP:journals/corr/abs-2409-12191}, Video-LLaVA\cite{DBLP:conf/emnlp/LinYZCNJ024}, LLaVA-Video\cite{DBLP:journals/corr/abs-2410-02713}, and InternVL2.5\cite{DBLP:journals/corr/abs-2412-05271}. We also compare VideoMiner with some similar frameworks, such as LifelongMemory\cite{DBLP:journals/corr/abs-2312-05269},
 VideoTree\cite{DBLP:journals/corr/abs-2405-19209}, LLoVi\cite{DBLP:conf/emnlp/0010LIWYBB24}, VideoAgent\cite{DBLP:conf/eccv/WangZZY24} and VideoAgent\cite{10.1007/978-3-031-72670-5_5}, mainly in terms of effectiveness and efficiency.
\vspace{-5mm}
\paragraph{Evaluations.}We evaluate all datasets under the multiple-choice QA and free-form generation settings. For multiple-choice QA, we utilize standard accuracy metrics. For free-form generation, we employ a GPT-assisted evaluation to assess the quality of the generated answers.
\definecolor{highlight}{RGB}{230, 230, 230}
\begin{table*}[t]
\centering
\caption{Performance Comparison on Short and Long Video Benchmarks}
\label{tab:performance}
\scriptsize
\setlength{\tabcolsep}{4pt}
\begin{tabular}{@{}cllcccccccccc@{}}
\toprule
\multicolumn{2}{c}{\multirow{3}*{\smash{\raisebox{-1.5ex}{\textbf{Method}}}}} 
& \multirow{3}*{\smash{\raisebox{-1.5ex}{\textbf{Base Model}}}}
& \multicolumn{9}{c}{\textbf{Longvideo Understanding Benchmark}} \\

\cmidrule(lr){4-12}
& & & \multicolumn{1}{c}{\textbf{EgoSchema}} & \multicolumn{3}{c}{\textbf{Video-MME}} & \multicolumn{4}{c}{\textbf{Longvideobench}} & \textbf{MLVU} \\

\cmidrule(lr){5-7} \cmidrule(lr){8-11}
& & & & \shortstack{Short} & \shortstack{Medium} & \shortstack{Long} & \shortstack{(8,15s]} & \shortstack{[15,60s]} & \shortstack{(180,600s]} & \shortstack{(900,3600s]} & \textbf{M-Avg} \\ 
\midrule

\rowcolor{highlight}
\multicolumn{12}{c}{\textbf{End-to-End Open-Source LVLMs}} \\
& Video-LLaVA & Vicuna-7B & 48.2 & 45.3 & 38.0 & 35.8 & 43.1 & 44.6 & 36.4 & 34.4 & 47.3 \\
& LLaVA-Video & Qwen2-7B & 60.2 & \textbf{72.0} & 56.0 & 49.3 & \textbf{69.8} & 68.0 & 54.1 & 45.5 & 62.1 \\
& InternVL2.5 & InternVL-2-8B & 60.0 & 60.0 & 51.2 & 50.6 & 69.3 & \textbf{70.9} & 52.9 & 46.4 & 59.2 \\
& Qwen2-VL & Qwen2-7B & 50.0 & 64.0 & 51.7 & 45.8 & 68.8 & 67.4 & 45.0 & 38.0 & 60.1 \\ 
\midrule
\rowcolor{highlight}
\multicolumn{12}{c}{\textbf{Existing Baselines}} \\
& LifelongMemory & GPT-4 & 64.1 & 60.1 & 52.7 & 46.6 & 61.8 & 58.5 & 50.3 & 42.0 & 53.9 \\
& VideoAgent \cite{DBLP:conf/eccv/WangZZY24} (ECCV24) & GPT-4 & 60.2 & 57.0 & 48.3 & 46.2 & 61.1 & 55.9 & 48.8 & 39.5 & 52.2 \\
& VideoAgent \cite{10.1007/978-3-031-72670-5_5} (ECCV24) & GPT-4 & 62.8 & 57.5 & 51.1 & 48.1 & 62.0 & 57.7 & 50.8 & 45.0 & 55.4 \\
& VideoAgent \cite{DBLP:conf/eccv/WangZZY24} (ECCV24) & Qwen-plus & 56.2 & 53.3 & 49.7 & 37.8 & 54.6 & 55.2 & 45.1 & 43.5 & 52.5 \\
& LLovi (EMNLP24) & Qwen-plus & 62.8 & 62.5 & 55.7 & 50.6 & 62.5 & 57.7 & 48.3 & 39.5 & 54.9 \\
& VideoTree (CVPR25) & Qwen-plus & 59.8 & 55.5 & 49.2 & 39.3 & 61.0 & 57.5 & 48.4 & 44.6 & 51.6 \\

& {\textbf{VideoMiner (Ours)}} & Qwen2-VL-7B & \textbf{\underline{66.2}} & \underline{65.6} & \textbf{\underline{57.5}} & \textbf{\underline{52.2}} & \underline{65.1} & \underline{64.7} & \textbf{\underline{58.6}} & \textbf{\underline{49.3}} & \textbf{\underline{65.1}} \\ 
\rowcolor{white}
\midrule
\end{tabular}
\vspace{1mm}
\parbox{\textwidth}{\footnotesize
\textbf{Bolded values} denote the highest score in each column across \emph{all methods}; 
\underline{underlined values} denote the highest score within the \emph{existing baselines}.
}
\vspace{-6mm}
\end{table*}

\subsection{Implementation Details}
For each video benchmark, we first extract a keyframe set with our approach and then sample 32 frames to match the uniformly sampled frames of the baselines. In our experiments, we mainly use Qwen2-VL-7B as the base model for VideoMiner. For other frameworks, we follow their official model setups and default settings. We test all methods on long video benchmarks, comparing their performance and time efficiency to evaluate how well VideoMiner works compared to existing approaches. 

\subsection{Main Results}
We conduct a comprehensive comparison of our method with 10 other baselines across 9 different sub-tasks within 4 well-known video understanding benchmarks. Specifically, we apply six long-video benchmark sub-tasks, achieving SOTA (state-of-the-art) performance in all long-video understanding tasks. Additionally, we maintain optimal performance in most short-video tasks compared to baselines with external structures.

\textbf{Obs.\ding{182}: As video length increases, the performance gap between our VideoMiner and the baselines gradually widens, demonstrating superior performance in long-video understanding tasks.} In long-video understanding tasks, end-to-end baseline methods are often hindered by large amounts of redundant information, while other baselines with external structures frequently lose significant temporal information. This issue is particularly pronounced in hierarchical baseline methods, which struggle to accurately select keyframes, resulting in suboptimal performance. In contrast, our approach leverages scene segmentation and clustering to maximally preserve temporal information. Furthermore, we employ reinforcement learning to train a policy model capable of self-directed decision-making, significantly enhancing its decision-making capacity. Consequently, our method effectively eliminates redundant information, improves the quality of selected keyframes, and strengthens the ability to understand long videos as shown in Table \ref{tab:performance}.
However, there is a certain performance gap compared to end-to-end methods among short video tasks. This is because the end-to-end methods and ours use different base models, and their models have been specifically trained and enhanced for video tasks. When the base model is the same, our plug-in architecture delivers a clear performance gain. Furthermore, our approach is primarily designed for long-video understanding, where keyframe selection is essential, while it is unnecessary for shorter videos. Nevertheless, our VideoMiner continues to outperform numerous baselines with external structures.

\begin{figure}[ht]
\begin{center}
\begin{minipage}{\columnwidth}
    \centering
   
    \begin{subfigure}[t]{0.49 \columnwidth}
        \includegraphics[width=\columnwidth]{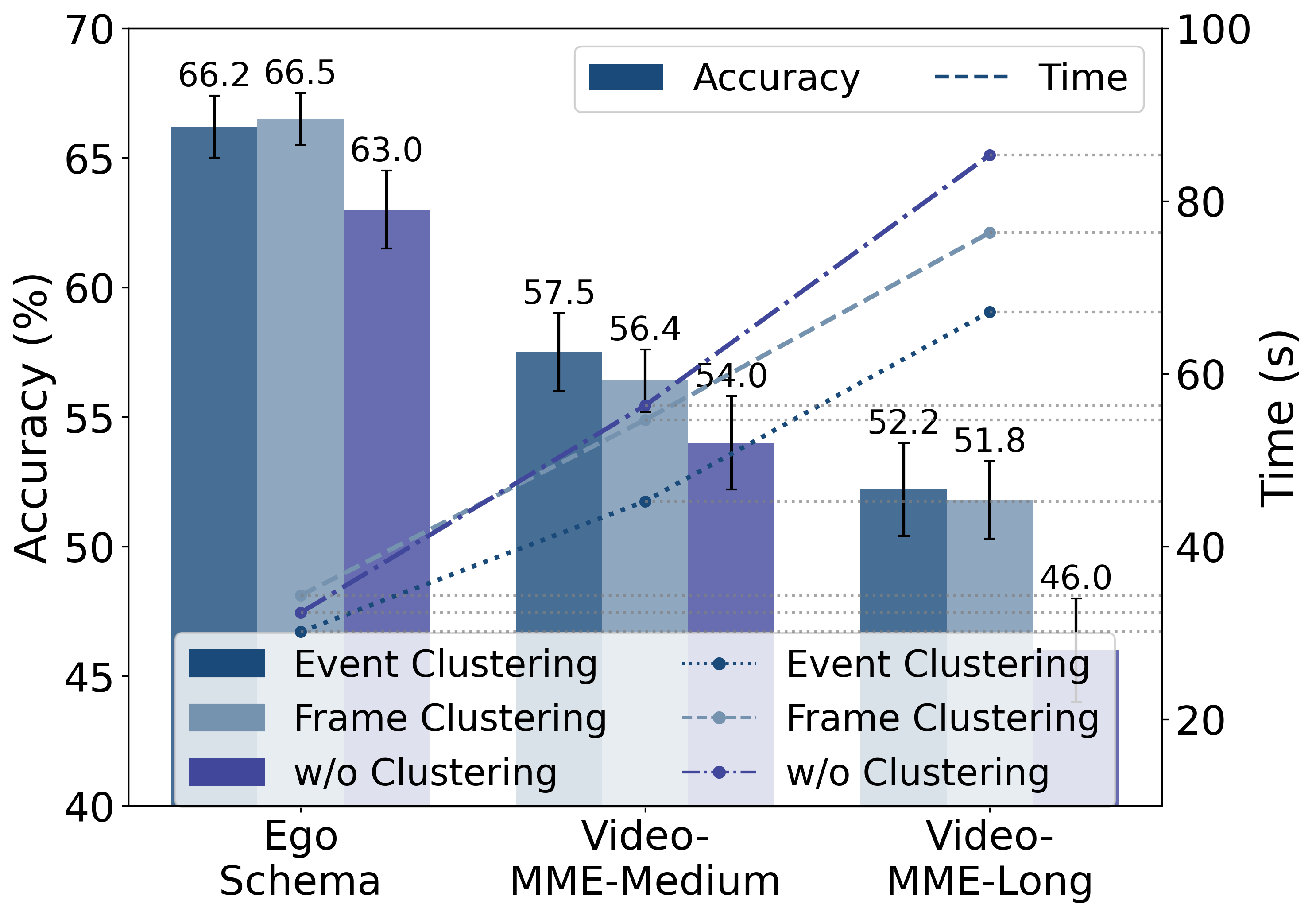}
        \caption{}
        \label{fig:subfig1_1}
    \end{subfigure}
    \hfill
    \begin{subfigure}[t]{0.49 \columnwidth}
        \includegraphics[width=\columnwidth]{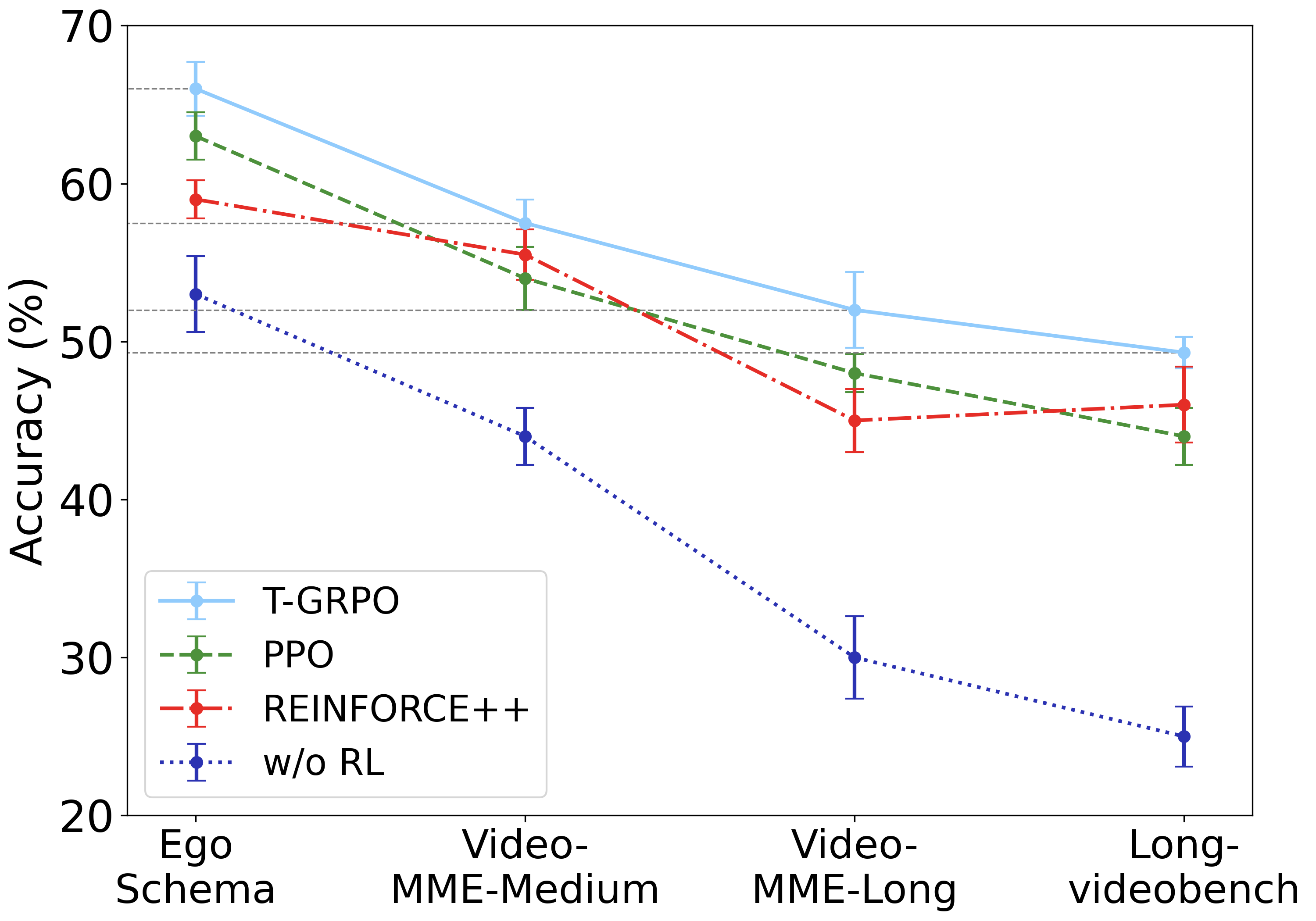}
        \caption{}
        \label{fig:subfig1_2}
    \end{subfigure}
    \hfill
\end{minipage}

\caption{Ablation study of clustering and reinforcement learning methods. (a) evaluates the impact of different clustering methods on accuracy and efficiency, while (b) analyzes the effect of various reinforcement learning approaches on accuracy.}
\label{fig:main1}
\end{center}
\vspace{-8mm}
\end{figure}
\subsection{Ablation Study}

\paragraph{Impact of cluster methods.}
We conducted ablation studies on four long-video benchmarks to evaluate the performance variations of our VideoMiner under three different settings: scene clustering, frame clustering, and without using any clustering method.  

\noindent
\textbf{Obs.\ding{183}: Compared to frame clustering, our proposed event clustering preserves richer temporal information and facilitates the efficient construction of the tree structure.}
As illustrated in Figure \ref{fig:subfig1_1}, event clustering achieves the shortest runtime and highest accuracy across most benchmarks. Additionally, clustering-based methods generally outperform non-clustering methods. This is because clustering methods significantly control the number of nodes at each layer through clustering, whereas non-clustering methods experience exponential growth in node numbers. Event clustering, in particular, retains more temporal information, allowing the policy model to make earlier and more precise decisions regarding node acceptance or deletion, thereby improving both effectiveness and efficiency.

\vspace{-5.5mm}
\paragraph{Impact of RL methods.}
We trained the policy model using different reinforcement learning algorithms on three long-video understanding benchmarks. Across all datasets, our proposed T-GRPO method consistently achieved the highest accuracy levels.

\noindent
\textbf{Obs.\ding{184}: Our T-GRPO introduces the tree-level reward design, significantly enhancing the inference capability of the policy model.} As shown in Figure \ref{fig:subfig1_2}, the untrained base model performs the worst across all benchmarks, and its performance deteriorates further as video length increases. RF methods without tree-level reward design show a significant improvement over the untrained baseline. Our T-GRPO method, by combining tree-level reward design, enables the policy model to take into account the impact of current decisions on future outcomes. This greatly enhances the inference capability of the policy model, ultimately leading to improved accuracy.

\subsection{Case Study}
\begin{figure}[htbp]
    \centering
    \includegraphics[width=\columnwidth,trim= 0 5 0 5]{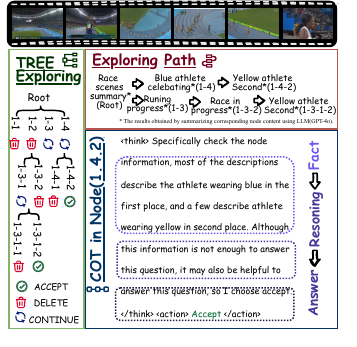} 
    \caption{Case study of the proposed VideoMiner. We present the tree node exploring path and the detailed reasoning process. Our proposed T-GRPO incentivizes the chain of thought of policy model, boosting reasoning ability of LLMs.}
    \label{fig:case}
    \vspace{-5mm}
\end{figure}

To visualize the procedure of our VideoMiner, we present a case study in Figure \ref{fig:case}. The input is a long video of a sports competition, with the question asking for the second athlete to cross the finish line. The video is first processed by VideoMiner, which segments, captions, and clusters the long video into a hierarchical tree structure. Then, the policy model trained with T-GRPO performs reasoning at each node. The tree exploring path is given in Figure \ref{fig:case}. Based on the node information, the policy model analyzes existing facts, and forms a reasoning chain to determine whether a frame qualifies as a key frame. This case demonstrates that T-GRPO encourages the policy model to generate responses with an extended reasoning chain style, significantly enhancing its inference capabilities. More case studies are given in Appendix B of the supplementary material.

\begin{figure}[ht]
\vspace{-2mm}
\begin{center}
\begin{minipage}{\columnwidth}
    \centering
   
    \begin{subfigure}[t]{0.49\columnwidth}
        \includegraphics[width=\columnwidth]{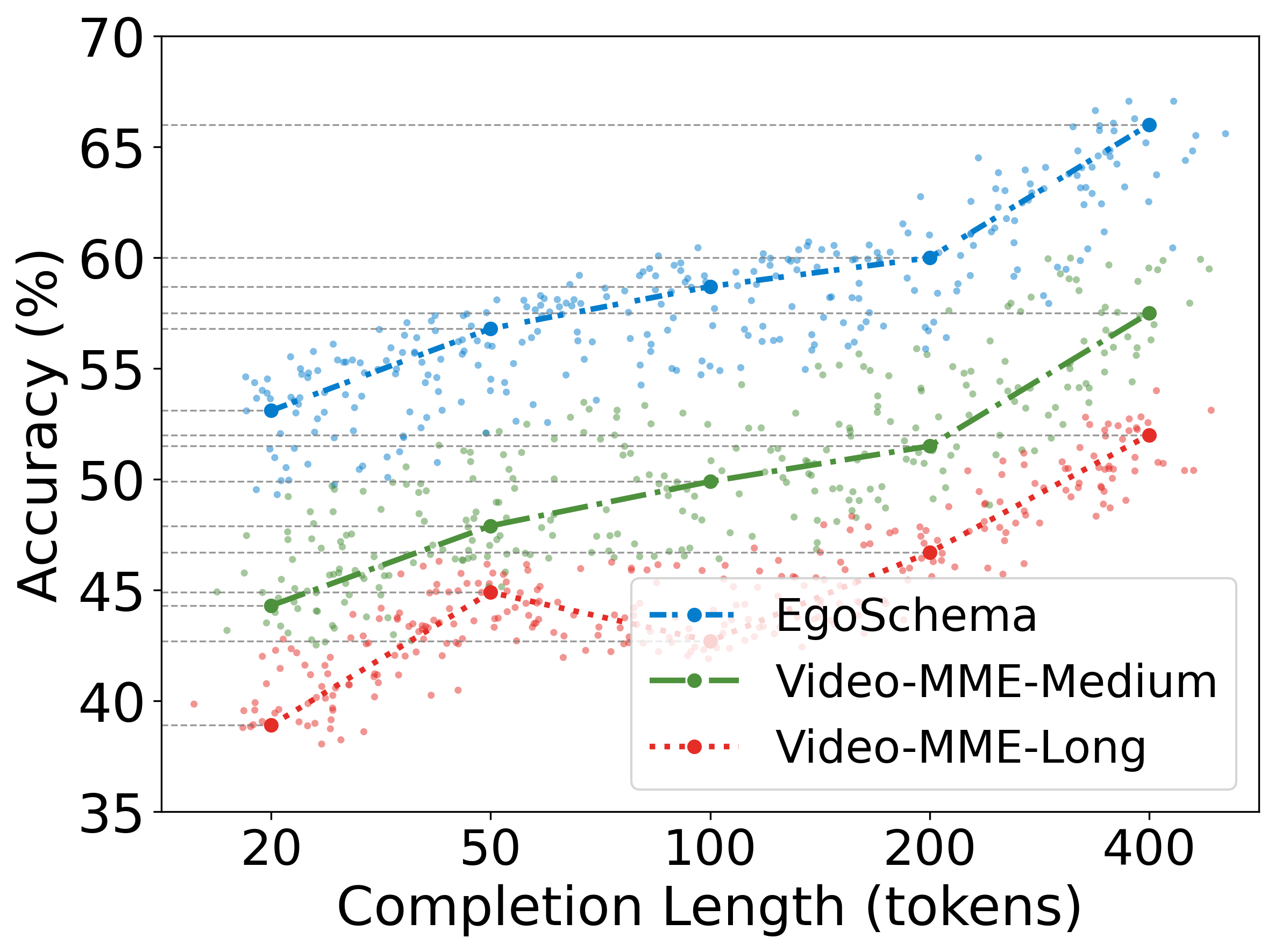}
        \caption{}
        \label{fig:subfig1}
    \end{subfigure}
    \hfill
    \begin{subfigure}[t]{0.49 \columnwidth}
        \includegraphics[width=\columnwidth]{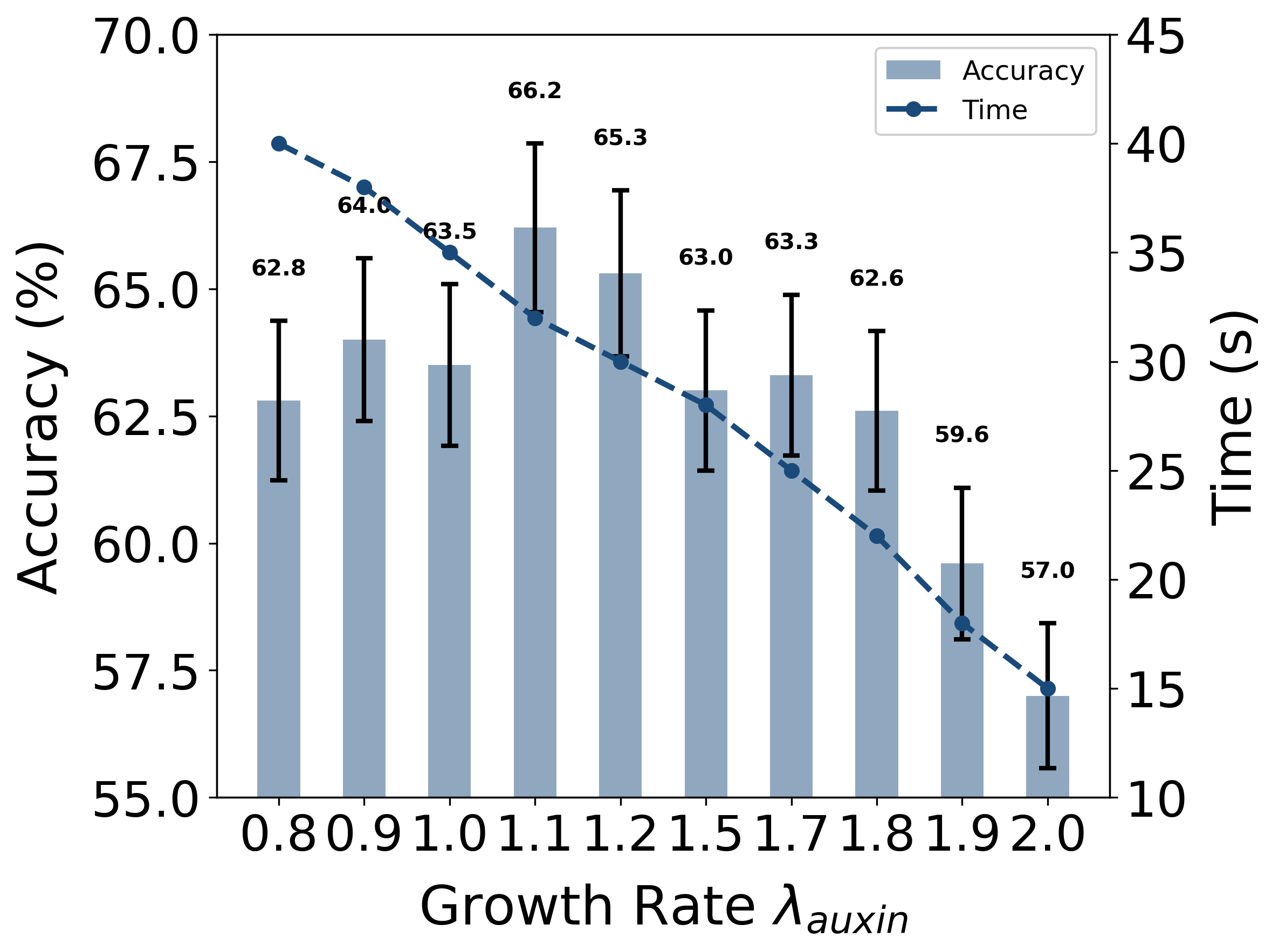}
        \caption{}
        \label{fig:subfig2}
    \end{subfigure}
    \hfill
\end{minipage}

\caption{(a) illustrates the impact of complement length on accuracy in the proposed T-GRPO framework, while (b) demonstrates how the tree growth rate \( \lambda_{auxin} \) in T-GRPO affects both the accuracy and efficiency of long video understanding tasks.}
\label{fig:main2}
\end{center}
\vspace{-8mm}
\end{figure}

\subsection{Discussion}
In this section, we will discuss some intriguing findings from our experiments and explore potential directions for future research. Our experiments revealed a correlation between the length of complement and the performance of VideoMiner. Additionally, we noted a balance between efficiency and performance resulting from growth rate \( \lambda_{auxin} \).
\vspace{-6mm}
\paragraph{Impact of Complement Length.}
We investigate the relationship between complement length and model performance by employing a Gaussian distribution-based length reward to monitor and select specific response length versions of the model, as depicted in Figure \ref{fig:subfig1}. Our findings across three benchmarks indicate a general trend: longer response lengths, or extended complement processes, lead to higher accuracy. Specifically, when the average number of output tokens increased to 400, accuracy improved by over 10\% compared to the initial 20 tokens. This enhancement occurs because the reinforcement learning process naturally induces chain-of-thought behaviors, significantly boosting the model's inferential capabilities. Consequently, this improves the model's ability to accurately identify key frames, thereby enhancing VideoMiner's performance.
\vspace{-5mm}
\paragraph{Influence of Growth Rate \( \lambda_{auxin} \).}
We control the policy model's action output tendencies by setting different ratios of action rewards, as shown in Figure \ref{fig:subfig2}. We define the ratio of the mean rewards for the ``accept'' and ``delete'' actions to the continue reward as the growth rate, reflecting the model's preference for early stopping actions (accept and delete) versus exploratory actions. Our observations show that a smaller growth rate leads the model to favor continuing actions, resulting in more thorough exploration with slower yet higher accuracy. As the growth rate increases, the model tends to output accept and delete actions earlier. Notably, when the growth rate is less than 1, the model may engage in aimless exploration, failing to discern useful information and ultimately compromising performance.

\section{Conclusion}

This paper presents VideoMiner, a novel long video understanding tree structure that adaptively ground key frames via the proposed T-GRPO. Our proposed VideoMiner, which iteratively segments, captions, and clusters long videos into a hierarchical tree structure, preserving temporal coherence from videos to events. To precisely locate key frames, we introduce T-GRPO, a tree-based reinforcement learning method that optimizes exploration within VideoMiner. Our approach achieves state-of-the-art performance in long video understanding tasks and reveals intriguing insights. Notably, T-GRPO encourages the spontaneous emergence of reasoning chains. Additionally, the tree growth auxin dynamically regulates expansion depth, balancing accuracy and efficiency.

\section{Acknowledgements}

We would like to express our sincere appreciation to Haolang Lu for designing and producing the high-quality figures presented in this paper. This work was supported by the National Natural Science Foundation of China under Grant 62471064.

{
    \small
    \bibliographystyle{ieeenat_fullname}
    \bibliography{ref}

\begin{thebibliography}{51}
\providecommand{\natexlab}[1]{#1}
\providecommand{\url}[1]{\texttt{#1}}
\expandafter\ifx\csname urlstyle\endcsname\relax
  \providecommand{\doi}[1]{doi: #1}\else
  \providecommand{\doi}{doi: \begingroup \urlstyle{rm}\Url}\fi

\bibitem[Aafaq et~al.(2019)Aafaq, Akhtar, Liu, Gilani, and Mian]{DBLP:conf/cvpr/AafaqALGM19}
Nayyer Aafaq, Naveed Akhtar, Wei Liu, Syed~Zulqarnain Gilani, and Ajmal Mian.
\newblock Spatio-temporal dynamics and semantic attribute enriched visual encoding for video captioning.
\newblock In \emph{{IEEE} Conference on Computer Vision and Pattern Recognition, {CVPR} 2019, Long Beach, CA, USA, June 16-20, 2019}, pages 12487--12496. Computer Vision Foundation / {IEEE}, 2019.

\bibitem[Buch et~al.(2022)Buch, Eyzaguirre, Gaidon, Wu, Fei{-}Fei, and Niebles]{DBLP:conf/cvpr/BuchEG00N22}
Shyamal Buch, Crist{\'{o}}bal Eyzaguirre, Adrien Gaidon, Jiajun Wu, Li Fei{-}Fei, and Juan~Carlos Niebles.
\newblock Revisiting the "video" in video-language understanding.
\newblock In \emph{{IEEE/CVF} Conference on Computer Vision and Pattern Recognition, {CVPR} 2022, New Orleans, LA, USA, June 18-24, 2022}, pages 2907--2917. {IEEE}, 2022.

\bibitem[Caffagni et~al.(2024)Caffagni, Cocchi, Barsellotti, Moratelli, Sarto, Baraldi, Cornia, and Cucchiara]{DBLP:conf/acl/CaffagniCBMS0CC24}
Davide Caffagni, Federico Cocchi, Luca Barsellotti, Nicholas Moratelli, Sara Sarto, Lorenzo Baraldi, Marcella Cornia, and Rita Cucchiara.
\newblock The revolution of multimodal large language models: {A} survey.
\newblock In \emph{Findings of the Association for Computational Linguistics, {ACL} 2024, Bangkok, Thailand and virtual meeting, August 11-16, 2024}, pages 13590--13618. Association for Computational Linguistics, 2024.

\bibitem[Chen et~al.(2024{\natexlab{a}})Chen, Lv, Wu, Lin, Song, Gao, Liu, Gao, Mao, and Shou]{DBLP:conf/cvpr/ChenLWLSGLGMS24}
Joya Chen, Zhaoyang Lv, Shiwei Wu, Kevin~Qinghong Lin, Chenan Song, Difei Gao, Jia{-}Wei Liu, Ziteng Gao, Dongxing Mao, and Mike~Zheng Shou.
\newblock Videollm-online: Online video large language model for streaming video.
\newblock In \emph{{IEEE/CVF} Conference on Computer Vision and Pattern Recognition, {CVPR} 2024, Seattle, WA, USA, June 16-22, 2024}, pages 18407--18418. {IEEE}, 2024{\natexlab{a}}.

\bibitem[Chen et~al.(2018)Chen, Wang, Zhang, and Huang]{DBLP:conf/eccv/ChenWZH18}
Yangyu Chen, Shuhui Wang, Weigang Zhang, and Qingming Huang.
\newblock Less is more: Picking informative frames for video captioning.
\newblock In \emph{Computer Vision - {ECCV} 2018 - 15th European Conference, Munich, Germany, September 8-14, 2018, Proceedings, Part {XIII}}, pages 367--384. Springer, 2018.

\bibitem[Chen et~al.(2024{\natexlab{b}})Chen, Wang, Cao, Liu, Gao, Cui, Zhu, Ye, Tian, and et~al]{DBLP:journals/corr/abs-2412-05271}
Zhe Chen, Weiyun Wang, Yue Cao, Yangzhou Liu, Zhangwei Gao, Erfei Cui, Jinguo Zhu, Shenglong Ye, Hao Tian, and et al.
\newblock Expanding performance boundaries of open-source multimodal models with model, data, and test-time scaling.
\newblock \emph{CoRR}, abs/2412.05271, 2024{\natexlab{b}}.

\bibitem[Dong et~al.(2019)Dong, Zhang, and Tan]{Dong_Zhang_Tan_2019}
Wenkai Dong, Zhaoxiang Zhang, and Tieniu Tan.
\newblock Attention-aware sampling via deep reinforcement learning for action recognition.
\newblock \emph{Proceedings of the AAAI Conference on Artificial Intelligence}, 33\penalty0 (01):\penalty0 8247--8254, 2019.

\bibitem[Doshi and Yilmaz(2020)]{DBLP:conf/cvpr/DoshiY20}
Keval Doshi and Yasin Yilmaz.
\newblock Continual learning for anomaly detection in surveillance videos.
\newblock In \emph{2020 {IEEE/CVF} Conference on Computer Vision and Pattern Recognition, {CVPR} Workshops 2020, Seattle, WA, USA, June 14-19, 2020}, pages 1025--1034. Computer Vision Foundation / {IEEE}, 2020.

\bibitem[Du et~al.(2024{\natexlab{a}})Du, Nan, Zhang, Xie, Xu, Fan, Cui, Tao, and Jiang]{du2024docmsu}
Hang Du, Guoshun Nan, Sicheng Zhang, Binzhu Xie, Junrui Xu, Hehe Fan, Qimei Cui, Xiaofeng Tao, and Xudong Jiang.
\newblock Docmsu: A comprehensive benchmark for document-level multimodal sarcasm understanding.
\newblock In \emph{Proceedings of the AAAI Conference on Artificial Intelligence}, pages 17933--17941, 2024{\natexlab{a}}.

\bibitem[Du et~al.(2024{\natexlab{b}})Du, Zhang, Xie, Nan, Zhang, Xu, Liu, Leng, Liu, Fan, Huang, Feng, Chen, Zhang, Li, Zhang, Chen, Cui, and Tao]{10654917}
Hang Du, Sicheng Zhang, Binzhu Xie, Guoshun Nan, Jiayang Zhang, Junrui Xu, Hangyu Liu, Sicong Leng, Jiangming Liu, Hehe Fan, Dajiu Huang, Jing Feng, Linli Chen, Can Zhang, Xuhuan Li, Hao Zhang, Jianhang Chen, Qimei Cui, and Xiaofeng Tao.
\newblock Uncovering what, why and how: A comprehensive benchmark for causation understanding of video anomaly.
\newblock In \emph{2024 IEEE/CVF Conference on Computer Vision and Pattern Recognition (CVPR)}, pages 18793--18803, 2024{\natexlab{b}}.

\bibitem[Fan et~al.(2025)Fan, Ma, Wu, Du, Li, Gao, and Li]{10.1007/978-3-031-72670-5_5}
Yue Fan, Xiaojian Ma, Rujie Wu, Yuntao Du, Jiaqi Li, Zhi Gao, and Qing Li.
\newblock Videoagent: A memory-augmented multimodal agent for video understanding.
\newblock In \emph{Computer Vision -- ECCV 2024}, pages 75--92, Cham, 2025. Springer Nature Switzerland.

\bibitem[Fu et~al.(2024)Fu, Dai, Luo, Li, Ren, Zhang, Wang, Zhou, Shen, Zhang, and et~al]{DBLP:journals/corr/abs-2405-21075}
Chaoyou Fu, Yuhan Dai, Yondong Luo, Lei Li, Shuhuai Ren, Renrui Zhang, Zihan Wang, Chenyu Zhou, Yunhang Shen, Mengdan Zhang, and et al.
\newblock Video-mme: The first-ever comprehensive evaluation benchmark of multi-modal llms in video analysis.
\newblock \emph{CoRR}, abs/2405.21075, 2024.

\bibitem[Gao et~al.(2022)Gao, Lei, Zeng, Song, Wang, and Shen]{DBLP:journals/tip/GaoLZSWS22}
Lianli Gao, Yu Lei, Pengpeng Zeng, Jingkuan Song, Meng Wang, and Heng~Tao Shen.
\newblock Hierarchical representation network with auxiliary tasks for video captioning and video question answering.
\newblock \emph{{IEEE} Trans. Image Process.}, 31:\penalty0 202--215, 2022.

\bibitem[He et~al.(2019)He, Zhao, Huang, Li, Liu, and Wen]{He_Zhao_Huang_Li_Liu_Wen_2019}
Dongliang He, Xiang Zhao, Jizhou Huang, Fu Li, Xiao Liu, and Shilei Wen.
\newblock Read, watch, and move: Reinforcement learning for temporally grounding natural language descriptions in videos.
\newblock \emph{Proceedings of the AAAI Conference on Artificial Intelligence}, 33\penalty0 (01):\penalty0 8393--8400, 2019.

\bibitem[Hua et~al.(2022)Hua, Wang, Rui, Shao, and Wang]{WOS:000761218500002}
Xia Hua, Xinqing Wang, Ting Rui, Faming Shao, and Dong Wang.
\newblock Adversarial reinforcement learning with object-scene relational graph for video captioning.
\newblock \emph{IEEE TRANSACTIONS ON IMAGE PROCESSING}, 31:\penalty0 2004--2016, 2022.

\bibitem[Islam et~al.(2024)Islam, Ho, Yang, Nagarajan, Torresani, and Bertasius]{DBLP:conf/cvpr/IslamHYNTB24}
Md~Mohaiminul Islam, Ngan Ho, Xitong Yang, Tushar Nagarajan, Lorenzo Torresani, and Gedas Bertasius.
\newblock Video recap: Recursive captioning of hour-long videos.
\newblock In \emph{{IEEE/CVF} Conference on Computer Vision and Pattern Recognition, {CVPR} 2024, Seattle, WA, USA, June 16-22, 2024}, pages 18198--18208. {IEEE}, 2024.

\bibitem[Lin et~al.(2024)Lin, Ye, Zhu, Cui, Ning, Jin, and Yuan]{DBLP:conf/emnlp/LinYZCNJ024}
Bin Lin, Yang Ye, Bin Zhu, Jiaxi Cui, Munan Ning, Peng Jin, and Li Yuan.
\newblock Video-llava: Learning united visual representation by alignment before projection.
\newblock In \emph{Proceedings of the 2024 Conference on Empirical Methods in Natural Language Processing, {EMNLP} 2024, Miami, FL, USA, November 12-16, 2024}, pages 5971--5984. Association for Computational Linguistics, 2024.

\bibitem[Lin et~al.(2022)Lin, Li, Lin, Ahmed, Gan, Liu, Lu, and Wang]{DBLP:conf/cvpr/LinLL0G0LW22}
Kevin Lin, Linjie Li, Chung{-}Ching Lin, Faisal Ahmed, Zhe Gan, Zicheng Liu, Yumao Lu, and Lijuan Wang.
\newblock Swinbert: End-to-end transformers with sparse attention for video captioning.
\newblock In \emph{{IEEE/CVF} Conference on Computer Vision and Pattern Recognition, {CVPR} 2022, New Orleans, LA, USA, June 18-24, 2022}, pages 17928--17937. {IEEE}, 2022.

\bibitem[Liu et~al.(2024)Liu, Li, Tang, Ge, Shan, and Li]{DBLP:conf/eccv/LiuLTGSL24}
Ruyang Liu, Chen Li, Haoran Tang, Yixiao Ge, Ying Shan, and Ge Li.
\newblock {ST-LLM:} large language models are effective temporal learners.
\newblock In \emph{Computer Vision - {ECCV} 2024 - 18th European Conference, Milan, Italy, September 29-October 4, 2024, Proceedings, Part {LVII}}, pages 1--18. Springer, 2024.

\bibitem[Liu et~al.(2022)Liu, Meng, Huang, Vlontzos, Rueckert, and Kainz]{WOS:000750373700004}
Tianrui Liu, Qingjie Meng, Jun-Jie Huang, Athanasios Vlontzos, Daniel Rueckert, and Bernhard Kainz.
\newblock Video summarization through reinforcement learning with a 3d spatio-temporal u-net.
\newblock \emph{IEEE TRANSACTIONS ON IMAGE PROCESSING}, 31:\penalty0 1573--1586, 2022.

\bibitem[Maaz et~al.(2024)Maaz, Rasheed, Khan, and Khan]{DBLP:conf/acl/0001RKK24}
Muhammad Maaz, Hanoona~Abdul Rasheed, Salman Khan, and Fahad Khan.
\newblock Video-chatgpt: Towards detailed video understanding via large vision and language models.
\newblock In \emph{Proceedings of the 62nd Annual Meeting of the Association for Computational Linguistics (Volume 1: Long Papers), {ACL} 2024, Bangkok, Thailand, August 11-16, 2024}, pages 12585--12602. Association for Computational Linguistics, 2024.

\bibitem[Mangalam et~al.(2023)Mangalam, Akshulakov, and Malik]{DBLP:conf/nips/MangalamAM23}
Karttikeya Mangalam, Raiymbek Akshulakov, and Jitendra Malik.
\newblock Egoschema: {A} diagnostic benchmark for very long-form video language understanding.
\newblock In \emph{Advances in Neural Information Processing Systems 36: Annual Conference on Neural Information Processing Systems 2023, NeurIPS 2023, New Orleans, LA, USA, December 10 - 16, 2023}, 2023.

\bibitem[Nan et~al.(2020)Nan, Guo, Sekuli{\'c}, and Lu]{nan2020reasoning}
Guoshun Nan, Zhijiang Guo, Ivan Sekuli{\'c}, and Wei Lu.
\newblock Reasoning with latent structure refinement for document-level relation extraction.
\newblock In \emph{Proceedings of the 58th Annual Meeting of the Association for Computational Linguistics}, pages 1546--1557, 2020.

\bibitem[Nan et~al.(2021)Nan, Qiao, Xiao, Liu, Leng, Zhang, and Lu]{DBLP:conf/cvpr/NanQXLLZL21}
Guoshun Nan, Rui Qiao, Yao Xiao, Jun Liu, Sicong Leng, Hao Zhang, and Wei Lu.
\newblock Interventional video grounding with dual contrastive learning.
\newblock In \emph{{IEEE} Conference on Computer Vision and Pattern Recognition, {CVPR} 2021, virtual, June 19-25, 2021}, pages 2765--2775. Computer Vision Foundation / {IEEE}, 2021.

\bibitem[Qian et~al.(2024)Qian, Dong, Zhang, Zang, Ding, Lin, and Wang]{DBLP:conf/nips/QianDZZDLW24}
Rui Qian, Xiaoyi Dong, Pan Zhang, Yuhang Zang, Shuangrui Ding, Dahua Lin, and Jiaqi Wang.
\newblock Streaming long video understanding with large language models.
\newblock In \emph{Advances in Neural Information Processing Systems 38: Annual Conference on Neural Information Processing Systems 2024, NeurIPS 2024, Vancouver, BC, Canada, December 10 - 15, 2024}, 2024.

\bibitem[Ren et~al.(2024)Ren, Yao, Li, Sun, and Hou]{DBLP:conf/cvpr/RenYL0H24}
Shuhuai Ren, Linli Yao, Shicheng Li, Xu Sun, and Lu Hou.
\newblock Timechat: {A} time-sensitive multimodal large language model for long video understanding.
\newblock In \emph{{IEEE/CVF} Conference on Computer Vision and Pattern Recognition, {CVPR} 2024, Seattle, WA, USA, June 16-22, 2024}, pages 14313--14323. {IEEE}, 2024.

\bibitem[Rong et~al.(2024)Rong, Leemann, Nguyen, Fiedler, Qian, Unhelkar, Seidel, Kasneci, and Kasneci]{DBLP:journals/pami/RongLNFQUSKK24}
Yao Rong, Tobias Leemann, Thai{-}trang Nguyen, Lisa Fiedler, Peizhu Qian, Vaibhav~V. Unhelkar, Tina Seidel, Gjergji Kasneci, and Enkelejda Kasneci.
\newblock Towards human-centered explainable {AI:} {A} survey of user studies for model explanations.
\newblock \emph{{IEEE} Trans. Pattern Anal. Mach. Intell.}, 46\penalty0 (4):\penalty0 2104--2122, 2024.

\bibitem[Shao et~al.(2024)Shao, Wang, Zhu, Xu, Song, Zhang, Li, Wu, and Guo]{DBLP:journals/corr/abs-2402-03300}
Zhihong Shao, Peiyi Wang, Qihao Zhu, Runxin Xu, Junxiao Song, Mingchuan Zhang, Y.~K. Li, Y. Wu, and Daya Guo.
\newblock Deepseekmath: Pushing the limits of mathematical reasoning in open language models.
\newblock \emph{CoRR}, abs/2402.03300, 2024.

\bibitem[Song et~al.(2024)Song, Chai, Wang, Zhang, Zhou, Wu, Chi, Guo, Ye, and et~al]{DBLP:conf/cvpr/SongCWZZWCG0ZLH24}
Enxin Song, Wenhao Chai, Guanhong Wang, Yucheng Zhang, Haoyang Zhou, Feiyang Wu, Haozhe Chi, Xun Guo, Tian Ye, and et al.
\newblock Moviechat: From dense token to sparse memory for long video understanding.
\newblock In \emph{{IEEE/CVF} Conference on Computer Vision and Pattern Recognition, {CVPR} 2024, Seattle, WA, USA, June 16-22, 2024}, pages 18221--18232. {IEEE}, 2024.

\bibitem[Tang et~al.(2018)Tang, Tian, Lu, Li, and Zhou]{WOS:000457843605049}
Yansong Tang, Yi Tian, Jiwen Lu, Peiyang Li, and Jie Zhou.
\newblock Deep progressive reinforcement learning for skeleton-based action recognition.
\newblock In \emph{2018 IEEE/CVF CONFERENCE ON COMPUTER VISION AND PATTERN RECOGNITION (CVPR)}, pages 5323--5332. IEEE; CVF; IEEE Comp Soc, 2018.
\newblock 31st IEEE/CVF Conference on Computer Vision and Pattern Recognition (CVPR), Salt Lake City, UT, JUN 18-23, 2018.

\bibitem[Wang et~al.(2024{\natexlab{a}})Wang, Bai, Tan, Wang, Fan, Bai, Chen, Liu, Wang, and et~al]{DBLP:journals/corr/abs-2409-12191}
Peng Wang, Shuai Bai, Sinan Tan, Shijie Wang, Zhihao Fan, Jinze Bai, Keqin Chen, Xuejing Liu, Jialin Wang, and et al.
\newblock Qwen2-vl: Enhancing vision-language model's perception of the world at any resolution.
\newblock \emph{CoRR}, abs/2409.12191, 2024{\natexlab{a}}.

\bibitem[Wang et~al.(2025)Wang, Cheng, and Wang]{WOS:001397798100003}
Rongrong Wang, Yuhu Cheng, and Xuesong Wang.
\newblock Constrained visual representation learning with bisimulation metrics for safe reinforcement learning.
\newblock \emph{IEEE TRANSACTIONS ON IMAGE PROCESSING}, 34:\penalty0 379--393, 2025.

\bibitem[Wang et~al.(2018)Wang, Chen, Wu, Wang, and Wang]{WOS:000457843604038}
Xin Wang, Wenhu Chen, Jiawei Wu, Yuan-Fang Wang, and William~Yang Wang.
\newblock Video captioning via hierarchical reinforcement learning.
\newblock In \emph{2018 IEEE/CVF CONFERENCE ON COMPUTER VISION AND PATTERN RECOGNITION (CVPR)}, pages 4213--4222. IEEE; CVF; IEEE Comp Soc, 2018.
\newblock 31st IEEE/CVF Conference on Computer Vision and Pattern Recognition (CVPR), Salt Lake City, UT, JUN 18-23, 2018.

\bibitem[Wang et~al.(2024{\natexlab{b}})Wang, Zhang, Zohar, and Yeung{-}Levy]{DBLP:conf/eccv/WangZZY24}
Xiaohan Wang, Yuhui Zhang, Orr Zohar, and Serena Yeung{-}Levy.
\newblock Videoagent: Long-form video understanding with large language model as agent.
\newblock In \emph{Computer Vision - {ECCV} 2024 - 18th European Conference, Milan, Italy, September 29-October 4, 2024, Proceedings, Part {LXXX}}, pages 58--76. Springer, 2024{\natexlab{b}}.

\bibitem[Wang et~al.(2023)Wang, Yang, and Ren]{DBLP:journals/corr/abs-2312-05269}
Ying Wang, Yanlai Yang, and Mengye Ren.
\newblock Lifelongmemory: Leveraging llms for answering queries in egocentric videos.
\newblock \emph{CoRR}, abs/2312.05269, 2023.

\bibitem[Wang et~al.(2024{\natexlab{c}})Wang, Yu, Stengel{-}Eskin, Yoon, Cheng, Bertasius, and Bansal]{DBLP:journals/corr/abs-2405-19209}
Ziyang Wang, Shoubin Yu, Elias Stengel{-}Eskin, Jaehong Yoon, Feng Cheng, Gedas Bertasius, and Mohit Bansal.
\newblock Videotree: Adaptive tree-based video representation for {LLM} reasoning on long videos.
\newblock \emph{CoRR}, abs/2405.19209, 2024{\natexlab{c}}.

\bibitem[Wu et~al.(2024{\natexlab{a}})Wu, Li, Chen, and Li]{DBLP:conf/nips/WuLCL24}
Haoning Wu, Dongxu Li, Bei Chen, and Junnan Li.
\newblock Longvideobench: {A} benchmark for long-context interleaved video-language understanding.
\newblock In \emph{Advances in Neural Information Processing Systems 38: Annual Conference on Neural Information Processing Systems 2024, NeurIPS 2024, Vancouver, BC, Canada, December 10 - 15, 2024}, 2024{\natexlab{a}}.

\bibitem[Wu et~al.(2020)Wu, Li, Liu, and Lin]{Wu_Li_Liu_Lin_2020}
Jie Wu, Guanbin Li, Si Liu, and Liang Lin.
\newblock {Tree-Structured Policy Based Progressive Reinforcement Learning for Temporally Language Grounding in Video}.
\newblock \emph{Proceedings of the AAAI Conference on Artificial Intelligence}, 34\penalty0 (07):\penalty0 12386--12393, 2020.

\bibitem[Wu et~al.(2024{\natexlab{b}})Wu, Fei, Qu, Ji, and Chua]{DBLP:conf/icml/Wu0Q0C24}
Shengqiong Wu, Hao Fei, Leigang Qu, Wei Ji, and Tat{-}Seng Chua.
\newblock Next-gpt: Any-to-any multimodal {LLM}.
\newblock In \emph{Forty-first International Conference on Machine Learning, {ICML} 2024, Vienna, Austria, July 21-27, 2024}. OpenReview.net, 2024{\natexlab{b}}.

\bibitem[Yang et~al.(2022)Yang, Miech, Sivic, Laptev, and Schmid]{DBLP:conf/cvpr/YangMSLS22}
Antoine Yang, Antoine Miech, Josef Sivic, Ivan Laptev, and Cordelia Schmid.
\newblock Tubedetr: Spatio-temporal video grounding with transformers.
\newblock In \emph{{IEEE/CVF} Conference on Computer Vision and Pattern Recognition, {CVPR} 2022, New Orleans, LA, USA, June 18-24, 2022}, pages 16421--16432. {IEEE}, 2022.

\bibitem[Ye et~al.(2022)Ye, Li, Qi, Wang, Huang, and Yang]{DBLP:conf/cvpr/YeLQWH022}
Hanhua Ye, Guorong Li, Yuankai Qi, Shuhui Wang, Qingming Huang, and Ming{-}Hsuan Yang.
\newblock Hierarchical modular network for video captioning.
\newblock In \emph{{IEEE/CVF} Conference on Computer Vision and Pattern Recognition, {CVPR} 2022, New Orleans, LA, USA, June 18-24, 2022}, pages 17918--17927. {IEEE}, 2022.

\bibitem[Yun et~al.(2018)Yun, Choi, Yoo, Yun, and Choi]{WOS:000432398300016}
Sangdoo Yun, Jongwon Choi, Youngjoon Yoo, Kimin Yun, and Jin~Young Choi.
\newblock Action-driven visual object tracking with deep reinforcement learning.
\newblock \emph{IEEE TRANSACTIONS ON NEURAL NETWORKS AND LEARNING SYSTEMS}, 29\penalty0 (6):\penalty0 2239--2252, 2018.

\bibitem[Zeng et~al.(2020)Zeng, Xu, Huang, Chen, Tan, and Gan]{DBLP:conf/cvpr/ZengXHCTG20}
Runhao Zeng, Haoming Xu, Wenbing Huang, Peihao Chen, Mingkui Tan, and Chuang Gan.
\newblock Dense regression network for video grounding.
\newblock In \emph{2020 {IEEE/CVF} Conference on Computer Vision and Pattern Recognition, {CVPR} 2020, Seattle, WA, USA, June 13-19, 2020}, pages 10284--10293. Computer Vision Foundation / {IEEE}, 2020.

\bibitem[Zhang et~al.(2024{\natexlab{a}})Zhang, Lu, Islam, Wang, Yu, Bansal, and Bertasius]{DBLP:conf/emnlp/0010LIWYBB24}
Ce Zhang, Taixi Lu, Md~Mohaiminul Islam, Ziyang Wang, Shoubin Yu, Mohit Bansal, and Gedas Bertasius.
\newblock A simple {LLM} framework for long-range video question-answering.
\newblock In \emph{Proceedings of the 2024 Conference on Empirical Methods in Natural Language Processing, {EMNLP} 2024, Miami, FL, USA, November 12-16, 2024}, pages 21715--21737. Association for Computational Linguistics, 2024{\natexlab{a}}.

\bibitem[Zhang et~al.(2023)Zhang, Li, and Bing]{DBLP:conf/emnlp/ZhangLB23}
Hang Zhang, Xin Li, and Lidong Bing.
\newblock Video-llama: An instruction-tuned audio-visual language model for video understanding.
\newblock In \emph{Proceedings of the 2023 Conference on Empirical Methods in Natural Language Processing, {EMNLP} 2023 - System Demonstrations, Singapore, December 6-10, 2023}, pages 543--553. Association for Computational Linguistics, 2023.

\bibitem[Zhang et~al.(2020)Zhang, Wang, Ma, and Liu]{WOS:000587912800009}
Wei Zhang, Bairui Wang, Lin Ma, and Wei Liu.
\newblock Reconstruct and represent video contents for captioning via reinforcement learning.
\newblock \emph{IEEE TRANSACTIONS ON PATTERN ANALYSIS AND MACHINE INTELLIGENCE}, 42\penalty0 (12):\penalty0 3088--3101, 2020.

\bibitem[Zhang et~al.(2024{\natexlab{b}})Zhang, Wu, Li, Li, Ma, Liu, and Li]{DBLP:journals/corr/abs-2410-02713}
Yuanhan Zhang, Jinming Wu, Wei Li, Bo Li, Zejun Ma, Ziwei Liu, and Chunyuan Li.
\newblock Video instruction tuning with synthetic data.
\newblock \emph{CoRR}, abs/2410.02713, 2024{\natexlab{b}}.

\bibitem[Zhao et~al.(2024)Zhao, Gundavarapu, Yuan, Zhou, Yan, Sun, Friedman, Qian, Weyand, Zhao, and et~al]{DBLP:conf/icml/0003GYZYSFQW0HS24}
Long Zhao, Nitesh~Bharadwaj Gundavarapu, Liangzhe Yuan, Hao Zhou, Shen Yan, Jennifer~J. Sun, Luke Friedman, Rui Qian, Tobias Weyand, Yue Zhao, and et al.
\newblock Videoprism: {A} foundational visual encoder for video understanding.
\newblock In \emph{Forty-first International Conference on Machine Learning, {ICML} 2024, Vienna, Austria, July 21-27, 2024}. OpenReview.net, 2024.

\bibitem[Zhou et~al.(2024)Zhou, Shu, Zhao, Wu, Xiao, Yang, Xiong, Zhang, Huang, and Liu]{DBLP:journals/corr/abs-2406-04264}
Junjie Zhou, Yan Shu, Bo Zhao, Boya Wu, Shitao Xiao, Xi Yang, Yongping Xiong, Bo Zhang, Tiejun Huang, and Zheng Liu.
\newblock {MLVU:} {A} comprehensive benchmark for multi-task long video understanding.
\newblock \emph{CoRR}, abs/2406.04264, 2024.

\bibitem[Zhou et~al.(2018)Zhou, Qiao, and Xiang]{Zhou_Qiao_Xiang_2018}
Kaiyang Zhou, Yu Qiao, and Tao Xiang.
\newblock Deep reinforcement learning for unsupervised video summarization with diversity-representativeness reward.
\newblock \emph{Proceedings of the AAAI Conference on Artificial Intelligence}, 32\penalty0 (1), 2018.

\bibitem[Zhu et~al.(2023)Zhu, Xia, Wu, Deng, Zhou, Qin, Liu, and Li]{WOS:000966336700001}
Jinhua Zhu, Yingce Xia, Lijun Wu, Jiajun Deng, Wengang Zhou, Tao Qin, Tie-Yan Liu, and Houqiang Li.
\newblock Masked contrastive representation learning for reinforcement learning.
\newblock \emph{IEEE TRANSACTIONS ON PATTERN ANALYSIS AND MACHINE INTELLIGENCE}, 45\penalty0 (3):\penalty0 3421--3433, 2023.

\end{thebibliography}
}

\end{document}